# BOLLWM: A REAL-WORLD DATASET FOR BOLLWORM PEST MONITORING FROM COTTON FIELDS IN INDIA

**Jerome White  Chandan Agrawal  Anmol Ojha**
**Apoorv Agnihotri  Makkunda Sharma  Jigar Doshi**
`agri-ai@wadhwaniai.org`
Wadhwani Institute for Artificial Intelligence

## ABSTRACT

This paper presents a dataset of agricultural pest images captured over five years by thousands of small holder farmers and farming extension workers across India. The dataset has been used to support a mobile application that relies on artificial intelligence to assist farmers with pest management decisions. Creation came from a mix of organized data collection, and from mobile application usage that was less controlled. This makes the dataset unique within the pest detection community, exhibiting a number of characteristics that place it closer to other non-agricultural objected detection datasets. This not only makes the dataset applicable to future pest management applications, it opens the door for a wide variety of other research agendas.

## 1 INTRODUCTION

Pest management plays a critical role for farmers around the world. This is especially true for smallholder farmers in low- and middle-income countries where failure to adhere to best practices can have devastating consequences (Guha & Das, 2022). Artificial intelligence can help, but requires the right kind of data to be effective—data that is not only sufficient in quantity for meaningful training, but also accurately representative of the deployment scenario.

This paper introduces an open source dataset that addresses that gap.[1] The dataset consists of images captured during the development and deployment of a mobile phone application (app) that assists farmers with pest management decisions (Dalmia et al., 2020). Collection has been ongoing since 2018, spanning several cotton farms and cotton seasons across India. The app provides recommendations for dealing with pink and American bollworms, two types of pests that have plagued the country's cotton farmers for decades (Mohan et al., 2017; Fand et al., 2019; Vonzun et al., 2019; Najork et al., 2021). Most images in the set consist of one or more of these pests, and come with annotated bounding boxes delineating their location.

As compared to other pest-specific datasets, our contribution is notable for its scale and its authentic representation of field conditions (Sections 2 and 3). It also has characteristics that make it comparable to other object detection datasets not in the agriculture domain (Section 4). Finally, as a result of collection-by-deployment, there are a number of images that do not contain pests (Section 5). Their inclusion not only opens the door for interesting research directions, but facilitates model development that is robust and practical.

## 2 RELATED WORK

There has been a steady stream of deep learning work focused on the detection of pests in digital images (Prajapati et al., 2016; Borowiec et al., 2022; Suto, 2022; Júnior & Rieder, 2020; Barbedo, 2020; Li et al., 2021; Verma, 2022; Mekha & Parthasarathy, 2022). A side effect of this effort has been the presentation of several pest-related datasets. Nam & Hung (2018), for example, use a database of hundreds of images from a confectionery factory; Xie et al. (2018) build a set of several thousand images from farms around China. The data presented in this work most closely resembles efforts to detect bollworms (Bakkay et al., 2018; Ding & Taylor, 2016) caught using pheromone traps (Sun et al., 2018; Ding & Taylor, 2016; Kalamatianos et al., 2018) installed on cotton fields (Sarangdhar & Pawar, 2017; Parikh et al., 2016; Revathi & Hemalatha, 2012). In many of these cases, the datasets are difficult to utilize because they are not available publicly (Nam & Hung, 2018; Ding & Taylor, 2016; Xie et al., 2018). That barrier aside, many were collected in controlled settings, either from internet searches (Liu et al., 2016; Ahmad et al., 2022; Wu et al., 2019), or using purpose-built trap imaging devices (Wang et al., 2020). In both cases, such sets are of limited value to our solution, as images from our users are often of more diverse quality than those found on the internet, or come with more noise than those taken with dedicated hardware.

---

[1]`https://github.com/WadhwaniAI/pest-management-opendata`

Recent work has sought to fill this gap by introducing publicly available datasets designed for building pest detection models. IP102 (Wu et al., 2019) is a collection of several thousand images across more than 100 species. The scale of this dataset—both with respect to images and pest classes—is comparable to what we present. Unlike our images however, which were collected by farmers and farm extension workers in cotton fields, their set was collated from images found on the internet. AgriPest (Wang et al., 2021) is a collection consisting of thousands of images across various crop and pest types. As with our dataset, their images are also collected in actual field conditions; however they do not contain the same crop or pest types we do. Pest24 (Wang et al., 2020) is a data set of almost 30,000 images containing almost 40 different pest species—the species in our dataset are a subset of theirs. Like our work, their dataset was collected on actual fields. However, because they use highly specialized monitoring devices, images are much more homogeneous that ours. We provide a closer comparison to IP102 and AgriPest in Section 4.

## 3 DATASET CREATION

Our dataset consists of images containing bollworms caught in pheromone traps on cotton farms across India. A pheromone trap is a class of pest traps in which a pheromone-emitting lure attracts male adult bollworms (Figure 3a). Each lure is designed to attract a specific species; for our collection those were pink and American bollworms, which we sometimes refer to as PBW and ABW, respectively. This section describes the two efforts that were undertaken to build our dataset.

### 3.1 DATA FROM DEDICATED COLLECTION

Data collection began with the explicit goal of building a pest detection model to support a mobile app for pest management. The intent was for farmers to take photos of bollworms caught in pheromone traps around a field. The app would then count relevant pests in the images and provide spray recommendations based on mappings between counts and agriculturally established best-practices. Having this solution clarity allowed us to design the data collection activity with protocols to build an appropriate set. To facilitate collection, partner organizations with expertise in cotton farming and having long-standing relationships with individual farmers were identified and engaged to assist. The organizations agreed that while making their regularly planned field visits, they would also collect images of trap catch.

Coordinated data collection began in early October 2018. A total of 26 farms were chosen in and around Wardha, a district in eastern Maharashtra (India). Each farm was outfitted with two pest traps specific to bollworms. Images were captured using ODK (Hartung et al., 2010). Forms were designed that allowed extension workers to first specify the lure type, then to take a photo of its contents. Workers were asked to empty the contents of the trap onto a plain white sheet of paper and take photos using reasonable photography techniques—to not engage the flash, and to ensure bollworms were in focus. They were further instructed to ensure that the photo was taken from a distance that minimized the space between the contents and the edge of the frame. Because photos were taken at a range of distances, the data set contains a corresponding range of pest scales.

Once Wardha's kharif season was finished, coordinated collection restarted in Nannilam, a town in Tamil Nadu approximately 1,300 kilometers south. Unlike Wardha, Nannilam grows cotton during the summer season, a time that is inactive for farmers in Wardha. This allowed us to continue collection relatively uninterrupted. Forty four farms were enrolled during this phase. Workers were requested to visit fields with a higher, more regular cadence than had been done during the previous season.

Table 1: Descriptive statistics of comparative datasets. See Sections 4 and A.1.1 for further details.

| Dataset | Images | | Boxes | | | | Classes | |
|---|---|---|---|---|---|---|---|---|
| | Total | Res. (MP) | Total | Per Image | Size | O/L | Total | Per Image |
| BOLLWM | 19 536 | 9.85±0.04 | 575 939 | 29.48±0.49 | 0.21% | 0.64 | 2 | 1.05±0.00 |
| BOLLWM/PBW | 16 176 | 9.62±0.04 | 536 653 | 33.18±0.58 | 0.15% | 0.65 | | |
| BOLLWM/ABW | 2 316 | 11.49±0.08 | 13 838 | 5.97±0.21 | 2.29% | 0.50 | | |
| IP102 | 15 177 | 0.33±0.01 | 17 839 | 1.18±0.01 | 37.18% | 0.05 | 97 | 1.00±0.00 |
| AgriPest | 28 169 | 3.33±0.02 | 144 798 | 5.14±0.07 | 0.28% | 0.18 | 13 | 1.00±0.00 |
| CARPK | 989 | 0.92±0.00 | 42 274 | 42.74±0.47 | 0.46% | 0.97 | 1 | |
| SKU-110K | 8 807 | 7.86±0.02 | 1 299 450 | 147.55±0.47 | 0.29% | 0.90 | 1 | |
| COCO/small | 63 911 | 0.27±0.00 | 371 655 | 5.82±0.02 | 0.30% | 0.45 | 80 | 2.05±0.01 |
| COCO/med. | 86 362 | 0.27±0.00 | 307 732 | 3.56±0.01 | 2.98% | 0.43 | 80 | 1.85±0.01 |

In both phases, quality assurance was maintained in two ways. First, regular feedback was provided through weekly conference calls and field visits. During conference calls, a sample subset of images were reviewed by the authors with the extension workers and appropriate feedback on picture quality given. Second, mobile chat groups consisting of extension workers, managers of the extension workers, and the authors of this paper were maintained. The group was used to quickly respond to general questions, provide feedback, and motivate extension workers to take high quality pictures. This was particularly important since data collection was not the primary focus of their field visits.

### 3.2 DATA FROM APPLICATION DEPLOYMENT

Using the data gathered from dedicated collection, object detection models were trained to recognize pink and American bollworms, facilitating development of our mobile app. During the 2019 kharif season, we deployed the app throughout Maharashtra, including to the same community of farmers and farming extension workers in Wardha with whom we interacted a year prior.

This phase marked a turning point in data collection: from that season onward images came exclusively from the app, which was deployed to a wider audience of users as the years went on. Prior to each deployment, users were educated on how to use the app, including instructions for how to empty traps and take photos such that the chances of model accuracy are highest. However, as usage scaled it became difficult to ensure the same level of protocol adherence as when collection was organized—the primary goal of the user base was pesticide decision making rather than positive feedback on image quality. This had noticeable implications on the range of data we received (Section 5), and on modeling techniques we pursued (Dalmia et al., 2020; White et al., 2022).

## 4 DATASET COMPARISONS

Characteristics of our dataset allow it to fit at the intersection of other popular object detection focused datasets, both within the pest detection space and otherwise. Our complete dataset (BOLLWM) consists of 47,851 images. The analysis in this section, presented in Table 1, is based on a subset of that total meant specifically for training. Similar training subsets—"train" and "validation"—were considered for third party datasets to keep comparisons fair. As such, numbers for datasets that are not ours may differ from their previously published results. In addition, our data was restricted to images that contained pests.

### 4.1 PEST DETECTION DATASETS

Rows IP102, AgriPest, and BOLLWM represent datasets devoted to pest detection. We have a similar number of images as IP102 (*images→total*) and a similar average pest size as AgriPest (*boxes→size*), likely because both we and AgriPest contain images taken in similar field conditions. IP102's larger pest size is likely an artifact of it being a curated set of internet images,

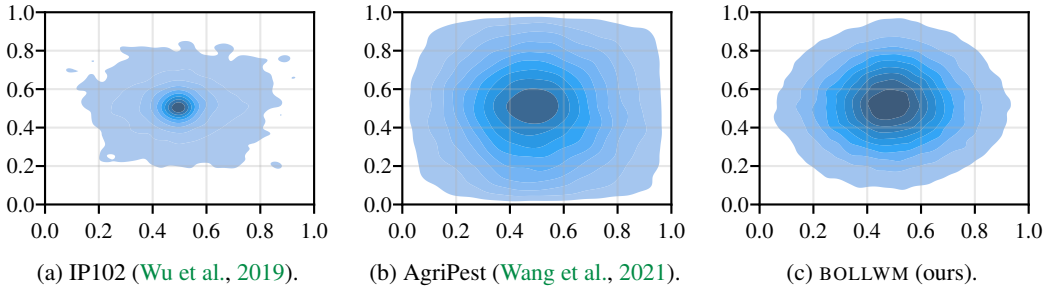

(a) IP102 (Wu et al., 2019).    (b) AgriPest (Wang et al., 2021).    (c) BOLLWM (ours).

Figure 1: Kernel density estimations of pest centroids. See Sections 5 and A.1.2 for further details.

which often fill the frame the crop of the pest. Our image resolution (*images→res.*) and total number of boxes (*boxes→total*) are much larger, in both cases by several multiples. We are also the only set with images that contain multiple classes (*classes→per image*).

Our dataset stands out for its pest density. The number of boxes per image (*boxes→per image*) and the number of images with at least one overlapping box (*boxes→O/L*) are proxies for this. The almost 30 boxes per image exhibited by our set is almost five times higher than the others. The standard error around that average is also much higher, meaning that there are some very tightly packed pest clusters in some of our images. Within the pest detection community, it has been found that this characteristic makes object detection more difficult (Wang et al., 2020).

Pest positions across the datasets also differ. Figure 1 presents density plots of bounding box centroids for each dataset. For each image box centroids were extracted and converted to positions relative to that image shape. Each panel in Figure 1 is the aggregation of these points by dataset. Lighter hues in a panel represent positions with fewer centroids; darker regions represent the opposite. In each dataset, pests tend to fall more heavily into the center of images. This is most acute in IP102. AgriPest tends to spread pests more equally around the image than other datasets. This is likely because many of their pests are found on the leaves and stems of plants, which tend to fill the frame. Our dataset has the widest distribution about the center.

### 4.2 GENERAL OBJECT DETECTION DATASETS

Some of the differentiating characteristics in our dataset are comparable to other datasets not dedicated to pest detection. Table 1 quantifies these differences through rows in the "general" domain. Those datasets include the Car Parking Lot Dataset (CARPK) (Hsieh et al., 2017), SKU-110K (Goldman et al., 2019), and Microsoft's Common Objects in Context (COCO) (Lin et al., 2014).

CARPK is a collection of car images taken by drones flown over parking lots. It was designed to facilitate research in object counting, but has also been used to explore methods on densely packed scenes (dos Santos de Arruda et al., 2022; Goldman et al., 2019). SKU-110K was designed specifically for detection in densely packed scenes. It is a collection of images of retail shelf displays, which require scores of intersecting bounding boxes to properly annotate. From Table 1, both datasets contain far more boxes per image than ours on average (*boxes→per image*); and images are much more likely to contain an overlapping pair (*boxes→O/L*). However, with respect to these metrics, our dataset is closer to these specialized corpora than to those in the pest detection space. Our dataset also contains more than one class, adding additional responsibility to models that want to count well in our domain.

COCO is a well established, multi-purpose dataset that is commonly used for object detection research. Although it is a more general purpose dataset, its popularity makes it convenient for couching comparisons in a well known context. In this case, we look specifically at its "small" and "medium" scale subsets, defined by the COCO maintainers as objects with area less-than 32 square pixels, and 96 square pixels, respectively. For either subset, the percentage of images containing overlapping boxes is lower than ours, but closer relatively than to the other pest detection sets. Our subset of ABW images has characteristics of both scale subsets: the number of boxes per image is in line with COCO-small, while the relative size of each box is comparable to COCO-medium.

## 5 IMAGE TYPES

Section 3 described the two phases of data collection. In the first, enumerators were explicitly tasked with taking photos that met our protocol. In the second collection became a side effect of a wider pest management application deployment. As mentioned, this had implications on the type of data added to our corpus.

Across phases there were changes to image quality, and to the diversity of pest configurations. Of note were the number of images that were submitted that did not have pests, and those that had pests, but on backgrounds other than white sheets of paper. Figure 2 visualizes this progression across all data splits. During the initial kharif and summer seasons—late 2018 and early 2019, respectively—almost all images contained pests; noted as the "positive" set in the figure. Once collection came from deployment, late 2019 onward, the number of "negative" images—images not containing pests—began to rise. Of the almost 48,000 images in our dataset, approximately 41 percent (19,558) are negatives. Section A.2 provides examples of all image types.

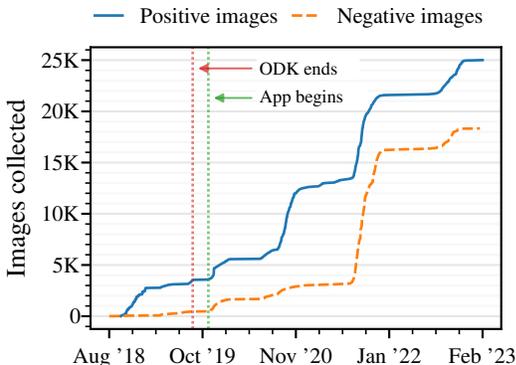

Figure 2: Progression of image collection.

Negative images are a natural occurrence for two reasons. The first is that there are genuine cases in which no pests are caught. There are also cases when users want to maintain their cadence of app usage, and knowingly submit no-pest images to do so. This set of negatives are generally white sheets of paper, as we request, that are either empty or filled with naturally occurring no-pest farm artifacts such as pieces of soil. The other reason for this negative set are users experimenting with the app. As with any application, it is natural for users to experiment before using it in its proper context to get a feel for its usage or to demonstrate the flow to show others. In such cases, images will not contain any evidence of pheromone trap contents, and instead be of common household or farm-related items.

Others looking to build their own pest management solutions based on uncontrolled user interaction will have to be resilient to this situation. Models that assume a strict domain are unlikely to perform well under such conditions. When our application is unable to correctly distinguish negative images from positive, for example, users lose trust and are hesitant to rely on the solution for such a critical component of their livelihoods. Inclusion of this negative set allows others to build models that are robust to a variety of input types.

## 6 CONCLUSION

Pest management is one of the most important responsibilities of a farmer. Good pest management involves consistent monitoring of pest populations, with appropriate action taken when those populations exceed a threshold (Kogan, 1998; Mancini et al., 2009; Rajashekhar et al., 2022). Although there are ongoing attempts to educate farmers on proper pest detection and intervention, successful programs require sizable on-ground support to scale and continued follow-up to be effective (Mancini et al., 2007). In the absence of such established, formalized, and accepted practices, pest detection can be ad hoc and pesticide usage excessive (Devi, 2010; Ranganathan, 2018); something that is not only expensive, but also dangerous to the environment and to human health (Donthi, 2021).

There is scope for artificial intelligence to mitigate this cycle. To do so, however, requires the right kind of data. This paper has presented a dataset that, due to the circumstances of its acquisition, represents a corpus that is closer to the ground reality than any other dataset in the open domain. The dataset is publicly available for researchers to develop computer vision techniques in a real-world setting, or for practitioners to explore new solutions to pest management that rely on artificial intelligence.

ACKNOWLEDGMENTS

The authors thank Balasaheb Dhame, Sonali Ghike, Rajesh Jain, Raghavendra Udupa, Srinivasa Rao, Kalyani Shastry, Sonu Vijan, and Dhruvin Vora for assistance at various stages of collection and curation.

We would also like to thank our program partners: Ambhuja Cement Foundation, Deshpande Foundation, Lupin, STL Garv, Syngenta, Tata Trust, and Welspun Foundation.

Portions of this work were made possible with support by the convening sponsor, "FAIR Forward—Artificial Intelligence for All," which is implemented by Deutsche Gesellschaft für Internationale Zusammenarbeit (GIZ) GmbH on behalf of the German Ministry for Economic Cooperation and Development (BMZ).

## A  APPENDIX

### A.1  EXTENDED CAPTIONS

#### A.1.1  DESCRIPTIVE STATISTICS

Table 1 provides descriptive statistics of our dataset alongside other object detection datasets:

**Image columns** Total images in the dataset along with average image resolution (image height multiplied by image width).

**Box columns** Total number of bounding boxes and the average number of bounding boxes per image. Size is the percentage of area each box consumes on average, relative to its parent image resolution. Overlap (*O/L*) represents the fraction of images in which at least one pair of boxes is overlapping.

**Class columns** Total number of bounding box classes, along with the average number of classes per image.

Slashes ("/") in dataset names denote data subsets. For example, $x/y$ denotes dataset $x$ restricted to subset $y$.

Uncertainty—values after each plus-minus—are standard errors about the mean. Uncertainty was omitted from size due to all values being much lower-than zero. All statistics are based on data from respective train and validation sets.

#### A.1.2  RELATIVE BOX POSITIONS

Figure 1 presents kernel density estimation plots of bounding box centroids. Centroids coordinates are normalized by the dimensions of their respective parent image. The $x$- and $y$-axis in each panel thus represents relative image width and height, respectively. Density estimates for each panel are independent; no normalization within or between datasets were taken. For this reason, color hues should not be compared across figures.

### A.2  IMAGE EXAMPLES

Figures 3, 4, and 5, and show examples of various image types in our dataset. Images have been scaled for brevity. Figure 3a is an example of a pheromone trap. Pests in figures 4, and 5 have been highlighted with their respective bounding boxes. Such boxes are not present in the actual raw data; they been added here to aid reader understanding.

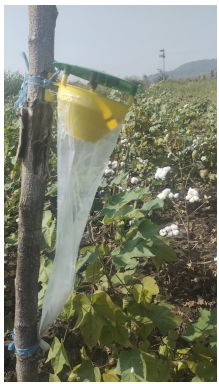
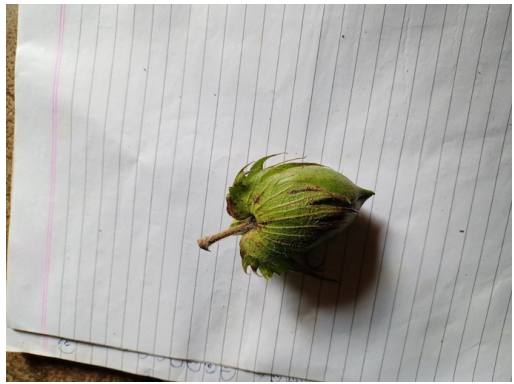

(a)                                    (b)

Figure 3: Examples of negative images in the dataset.

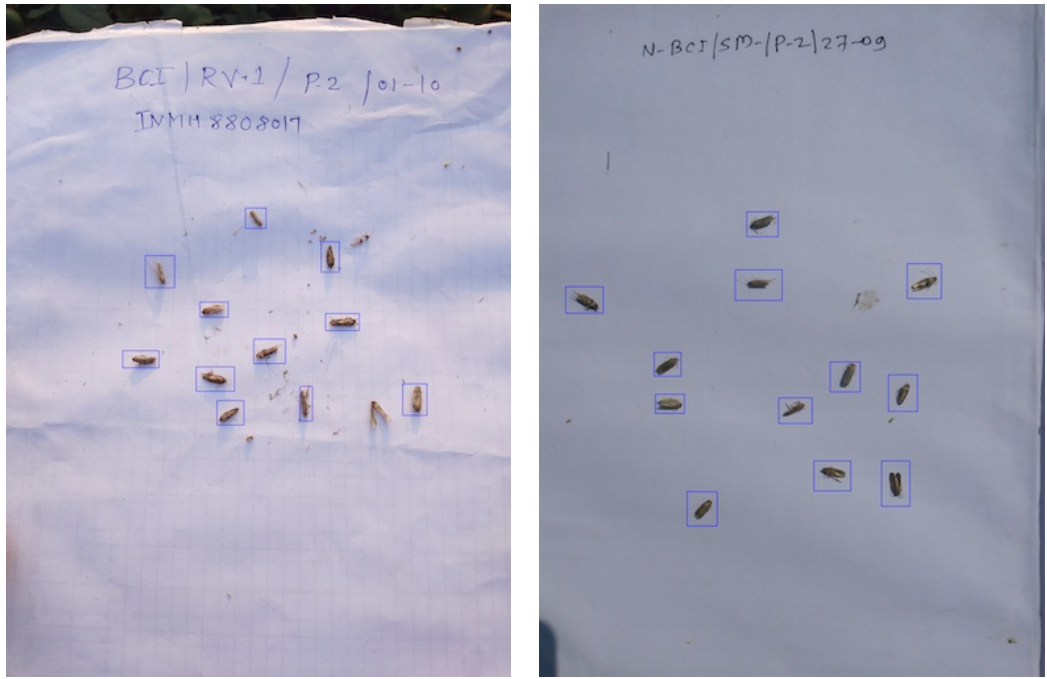

Figure 4: Examples of PBW images in the dataset.

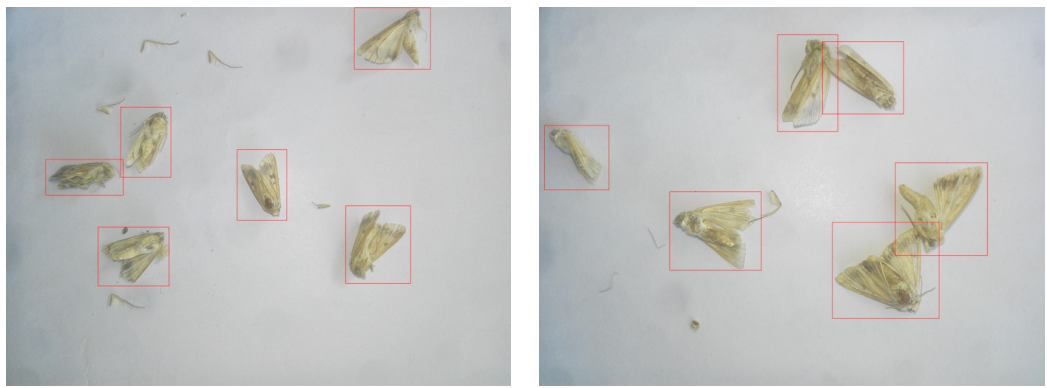

Figure 5: Examples of ABW images in the dataset.